\documentclass[journal]{IEEEtran}
\usepackage{amsfonts,amssymb}
\usepackage{amsmath}
\usepackage{adjustbox}
\usepackage{multirow}
\usepackage{subfigure}
\usepackage{algorithm}
\usepackage{algorithmic}

\ifCLASSINFOpdf
\else
\fi
\begin{document}
%
\title{Weakly Supervised Regional and Temporal Learning for Facial Action Unit Recognition}
%
%
%


\author{Jingwei~Yan$^*$,
        Jingjing~Wang$^*$,
        Qiang~Li,
        Chunmao~Wang,
        and~Shiliang~Pu$^\dag$
\IEEEcompsocitemizethanks{
\IEEEcompsocthanksitem J. Yan, J. Wang, Q. Li, C. Wang, and S. Pu are with Hikvision Research Institute, Hangzhou, 310051, China. E-mail: \{yanjingwei, wangjingjing9, liqiang23, wangchunmao, pushiliang.hri\}@hikvision.com.
\IEEEcompsocthanksitem $^*$ Equal contribution. 
\IEEEcompsocthanksitem $^\dag$ Corresponding author. 
}
}

%
%

\markboth{Journal of \LaTeX\ Class Files,~Vol.~14, No.~8, August~2015}%
{Shell \MakeLowercase{\textit{et al.}}: Bare Demo of IEEEtran.cls for IEEE Journals}
%



\maketitle

\begin{abstract}
Automatic facial action unit (AU) recognition is a challenging task due to the scarcity of manual annotations. To alleviate this problem, a large amount of efforts has been dedicated to exploiting various weakly supervised methods which leverage numerous unlabeled data. However, many aspects with regard to some unique properties of AUs, such as the regional and relational characteristics, are not sufficiently explored in previous works. Motivated by this, we take the AU properties into consideration and propose two auxiliary AU related tasks to bridge the gap between limited annotations and the model performance in a self-supervised manner via the unlabeled data. Specifically, to enhance the discrimination of regional features with AU relation embedding, we design a task of RoI inpainting to recover the randomly cropped AU patches. Meanwhile, a single image based optical flow estimation task is proposed to leverage the dynamic change of facial muscles and encode the motion information into the global feature representation. Based on these two self-supervised auxiliary tasks, local features, mutual relation and motion cues of AUs are better captured in the backbone network. Furthermore, by incorporating semi-supervised learning, we propose an end-to-end trainable framework named weakly supervised regional and temporal learning (WSRTL) for AU recognition. Extensive experiments on BP4D and DISFA demonstrate the superiority of our method and new state-of-the-art performances are achieved.
\end{abstract}


%
\IEEEpeerreviewmaketitle

\section{Introduction}
%
%
%
%
\IEEEPARstart{F}{acial}  action units describe a series of muscle movements on human faces precisely~\cite{ekman1997face}. Based on the combination of AUs, most facial expressions can be analyzed and modeled in an objective and explicit way. Nowadays, automatic AU recognition has drawn growing attention in the affective computing and computer vision communities and has been widely applied in human-computer interaction, fatigue monitoring, deception detection, etc.

The majority of existing AU recognition models are trained in a fully supervised manner which requires a correct manual annotation for each sample. However, the annotation of AUs not only needs facial action coding expertise, but also is very time-consuming, which results in limited data with reliable annotations. On the contrary, there is a large amount of easily accessible facial expression images or videos on the Internet which contain a diverse variety of AUs. Thus it is relatively convenient to build a large database without AU annotations from the Internet, such as Emotionet~\cite{fabian2016emotionet} and Affectnet~\cite{mollahosseini2017affectnet}, which both consist of more than one million facial images. For the purpose of making full use of the unlabeled data, many efforts were dedicated to exploring better feature presentation based on semi-supervised or unsupervised learning methods~\cite{peng2018weakly,niu2019multi,peng2019dual}. Inspired by these works, in this paper we will delve into the unlabeled data and learn discriminative feature representation for AU recognition from the aspect of both self-supervised auxiliary task learning and semi-supervised learning.

\begin{figure}[tbp]
	\begin{center}
		\includegraphics[width=\columnwidth]{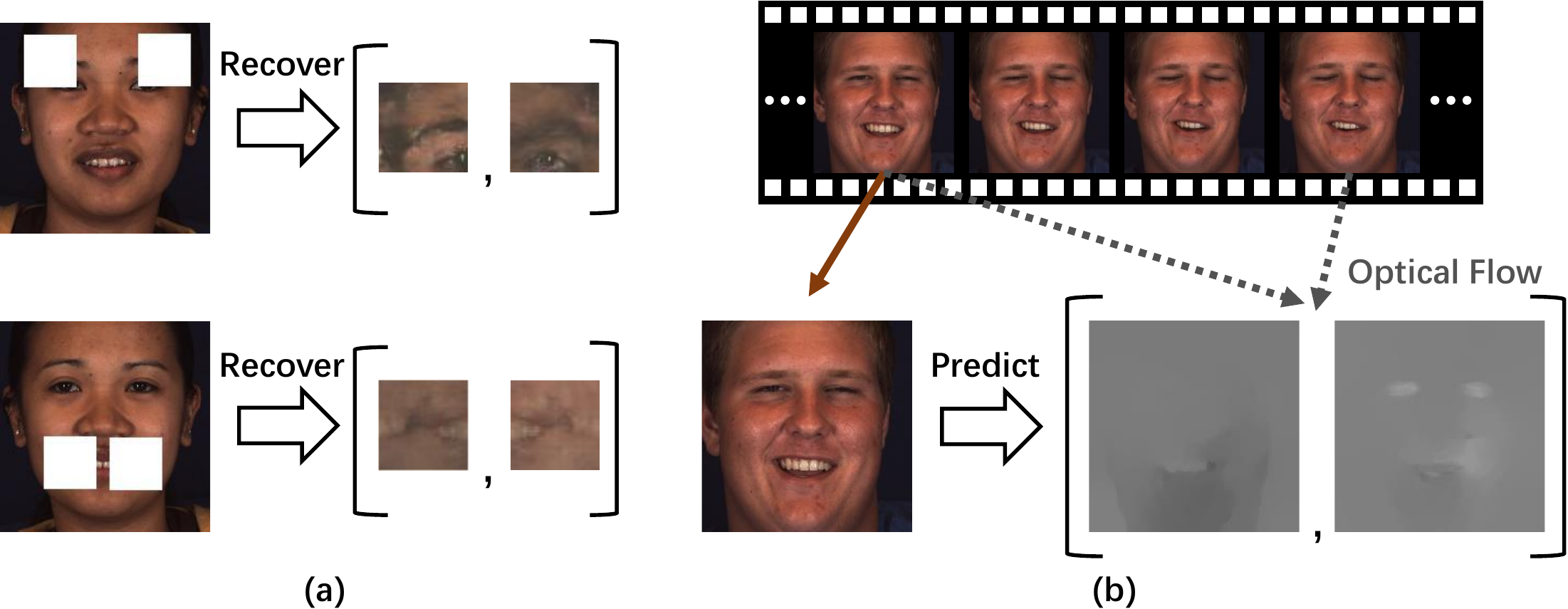}
	\end{center}
	\caption{Two proposed self-supervised auxiliary tasks. (a) RoI inpainting, which aims to recover the cropped AU RoIs via regional features embedded with AU relations. (b) Optical flow estimation, whose purpose is to explore temporal motion information of facial muscles from one static image.}
	\label{intro}
\end{figure}

Recently self-supervised learning has shown great potential in learning discriminative features from the unlabeled data via various pretext tasks~\cite{kolesnikov2019revisiting}. For the task of AU recognition, the limited works in this promising topic mainly focused on the aspect of global feature learning, which is in accordance with some previous self-supervised learning approaches for image recognition. Li et al.~\cite{li2019self} designed the task of reconstructing one facial frame in a video sequence in order to learn features for the movement of facial muscles. Lu et al.~\cite{lu2020self} leveraged the temporal consistency to learn feature representation by a pretext task similar to examplar~\cite{dosovitskiy2014discriminative}. In both methods, the self-supervised task was conducted on the unlabeled video sequences and only global features for facial frame were explored with the proposed pretext tasks. As a result, some unique properties of AUs, which are often leveraged to obtain discriminative AU features in fully or weakly supervised methods, such as the locality of AU and relationship between AUs, are ignored in these tasks. This will lead to a nonnegligible gap between the existing self-supervised tasks and AU recognition. For the purpose of bridging this gap, we propose a regional and temporal based auxiliary task learning method to incorporate regional and temporal feature learning, together with AU relation embedding simultaneously in one unified framework via two novel self-supervised tasks.

As each AU is corresponding to one or a small group of facial muscles, discriminative regional features are crucial for AU recognition. Individual convolutional layers are commonly applied on the region of interest (RoI) for different facial structures or textures. Meanwhile, the strong relationship between AUs is often leveraged to boost the recognition performance. Different from previous graph propagation methods which were based on the fixed prior AU relation graph, here a transformer~\cite{vaswani2017attention} is employed to encode the AU relation adaptively. By means of the self-attention mechanism in the transformer, relevant AU RoI features are aggregated to improve the representation capacity. However, the AU relation learned from limited labeled data may be flawed. In order to acquire accurate AU relation and enhance the representation ability of regional features through huge amounts of unlabeled data, a novel RoI inpainting (RoII) self-supervised task is designed. As shown in Figure~\ref{intro}(a), RoIs from the symmetrical parts of a random AU are cropped from the original image and then recovered based on the representative features of the copped patches. To obtain such features, other RoI features of the intact regions are leveraged to fuse together with the cropped ones via the learned AU relationship.

For some AUs like AU7 (lid tightener) and AU24 (lip presser), it is ambiguous to recognize them with the static facial image alone, especially when the intensity is weak, as the difference between low-intensity AUs and a neutral face can hardly be modeled via a single image. Therefore, it is beneficial to take the motion information of facial muscles into consideration. However, a sequence of frames is necessary to extract the temporal features conventionally and the computational costs of current multi-frame based AU recognition methods, such as 3D-CNN~\cite{tran2015learning} and CNN-LSTM~\cite{chu2016modeling}, are much heavier than image based models. To overcome the obstacle, given that the optical flow extracted from two frames naturally depicts such temporal change, we propose a self-supervised task of image based optical flow estimation (OFE), which is shown in Figure~\ref{intro}(b). The optical flow between two frames is extracted and served as the supervisory signal. Global features of the image are utilized to predict the optical flow so that the backbone network is forced to explore the muscle motions from the static image. Meanwhile, as the movements of the facial muscles only occur in certain regions caused by AU activation, the optical flow for supervision is sparse and only contains values in corresponding local facial parts. Therefore, another potential benefit of OFE is that the model will learn to focus on the important local regions automatically.

By integrating the two proposed auxiliary tasks, three crucial components for AU recognition, i.e., regional and temporal feature learning, together with AU relation encoding, are unified in one framework for the first time. Moreover, different from previous self-supervised learning methods which were conducted in two separate steps, i.e., training the feature extractor by performing the self-supervised task on unlabeled data and then training a linear classifier based on the labeled AU data, we incorporate the self-supervised tasks as auxiliary tasks which are trained with the AU recognition task simultaneously. Furthermore, a semi-supervised learning approach is integrated in to maximize the utilization of unlabeled data from a different aspect and form the final framework of weakly supervised regional and temporal learning (WSRTL) for AU recognition. The entire framework is efficient to train as it is end-to-end trainable and can achieve much better recognition performances compared to traditional self- or semi-supervised methods.

In summary, the major contributions of this paper are as follows.

\begin{itemize}
\item It is the first time that self- and semi-supervised learning are unified in one framework for AU recognition. The framework is end-to-end trainable and does not increase computational cost during inference.
\item Two novel AU specific auxiliary tasks are proposed to incorporate regional, temporal and AU relation learning in a self-supervised manner.
\item Transformer is employed to encode the AU relationship adaptively via the self-attention mechanism in a data driven fashion.
\item Based on the integrated WSRTL framework, state-of-the-art performances are achieved on two benchmark databases and the principle of designing self-supervised tasks will shed light on the development of self-supervised learning based AU recognition.
\end{itemize}

\section{Related Work}
\subsection{Action Unit Recognition}
Many efforts have been dedicated to AU recognition for the past decades. For the purpose of performance improvement, discriminative regional feature learning, AU relationship embedding and temporal feature learning are fully explored in supervised methods.

In order to capture regional features for different facial structures, Zhao et al.~\cite{zhao2016deep} proposed a region layer which divides the feature maps into identical regions, then independent convolutions were applied to each region. Niu et al.~\cite{niu2019local} proposed the LP-Net to model the person-specific shape information and explore the relationship among local regions simultaneously. Corneanu et al.~\cite{corneanu2018deep} and Zhang et al.~\cite{zhang2019context} cropped several relatively large regions from the original facial image and employed individual CNNs on them. Li et al.~\cite{li2017action} defined AU centers based on the rough position relationship between AUs and landmarks, and proposed a region of interest network to crop regions around these AU centers. This landmark based region localization method was followed by~\cite{shao2018deep,li2019semantic}.

The relationship between AUs has been exploited to boost the recognition performance. Corneanu et al.~\cite{corneanu2018deep} proposed to capture AU correlations by deep structured inference network which was applied on the AU predictions in a recurrent manner. Li et al.~\cite{li2019semantic} proposed a semantic relationship embedded representation learning (SRERL) framework to embed the relation knowledge in regional features with gated graph neural network. Graph convolutional network (GCN)~\cite{kipf2017semi} was also utilized to encode the prior AU relationship information into feature representation \cite{niu2019multi,liu2020relation,song2021uncertain,yan2021multi}. The graph propagation based methods often utilize a fixed AU relation graph to embed the relationship which is not suitable for some facial samples. In our framework, a transformer is leveraged due to the flexibility and the data driven manner and a self-supervised task is proposed to enhance the AU relation learning.

From the temporal aspect, an action unit essentially describes the motion of facial muscles. Thus it is helpful to leverage the temporal features of facial textures. To capture such dynamic change information, Chu et al.~\cite{chu2016modeling} and Jaiswal et al.~\cite{jaiswal2016deep} combined CNN and LSTM to model the AU sequence spatially and temporally. Similarly, Li et al.~\cite{li2017action} employed LSTM to fuse the temporal features of AU RoIs and improved the model performance greatly compared to the model without using temporal information. However, these methods are performed on the frame sequence and cannot be adapted to the image based models. Meanwhile, the complex network structures for temporal feature learning make the calculation inefficient during inference process. In our framework, a self-supervised auxiliary task is proposed to enhance the temporal representation capacity during training with a single static image, thus it does not increase the computational cost for AU recognition.

Some efforts were also made to integrate auxiliary tasks with AU recognition. Most of the approaches leveraged the recognition of facial expression or other related facial attributes which still required for corresponding manual annotations~\cite{yang2016multiple,chang2017fatauva,wang2017expression}. Shao et al.~\cite{shao2018deep} integrated the facial alignment task with AU recognition to learn better local features, while in our method, the two auxiliary tasks are designed to model AU characteristics from three crucial components via the unlabeled data in a self-supervised manner, which is more effective and does not need additional annotations.

\subsection{Self-Supervised Learning}
Self-supervised learning methods first deduce the ground truth labels or other supervisory signals from the designed pretext task itself. Then the ground truth is leveraged to supervise the training of the feature learning model. Through the designed self-supervised task, generic features are learned for the downstream task. For AU recognition, current works are mainly motivated by the temporal movement of facial muscles. Wiles et al.~\cite{koepke2018self} proposed FAb-Net to generate from the source frame to target frame by low-dimensional attribute embedding. Li et al.~\cite{li2019self} also utilized the transformation between the source and target frames and furthermore proposed the twin-cycle autoencoder to obtain dedicated AU features without head motions. Inspired by the intrinsic temporal consistency in videos, Lu et al.~\cite{lu2020self} used a triplet-based ranking approach to rank the frames. These methods mainly focus on discriminative global feature learning and ignore some unique properties of AUs. Regional feature leaning and AU relation embedding are not taken into consideration. Meanwhile, these self-supervised tasks are performed on video samples and cannot be adopted to tremendous unlabeled facial images.

\begin{figure}
	\begin{center}
		\includegraphics[width=1.0\columnwidth]{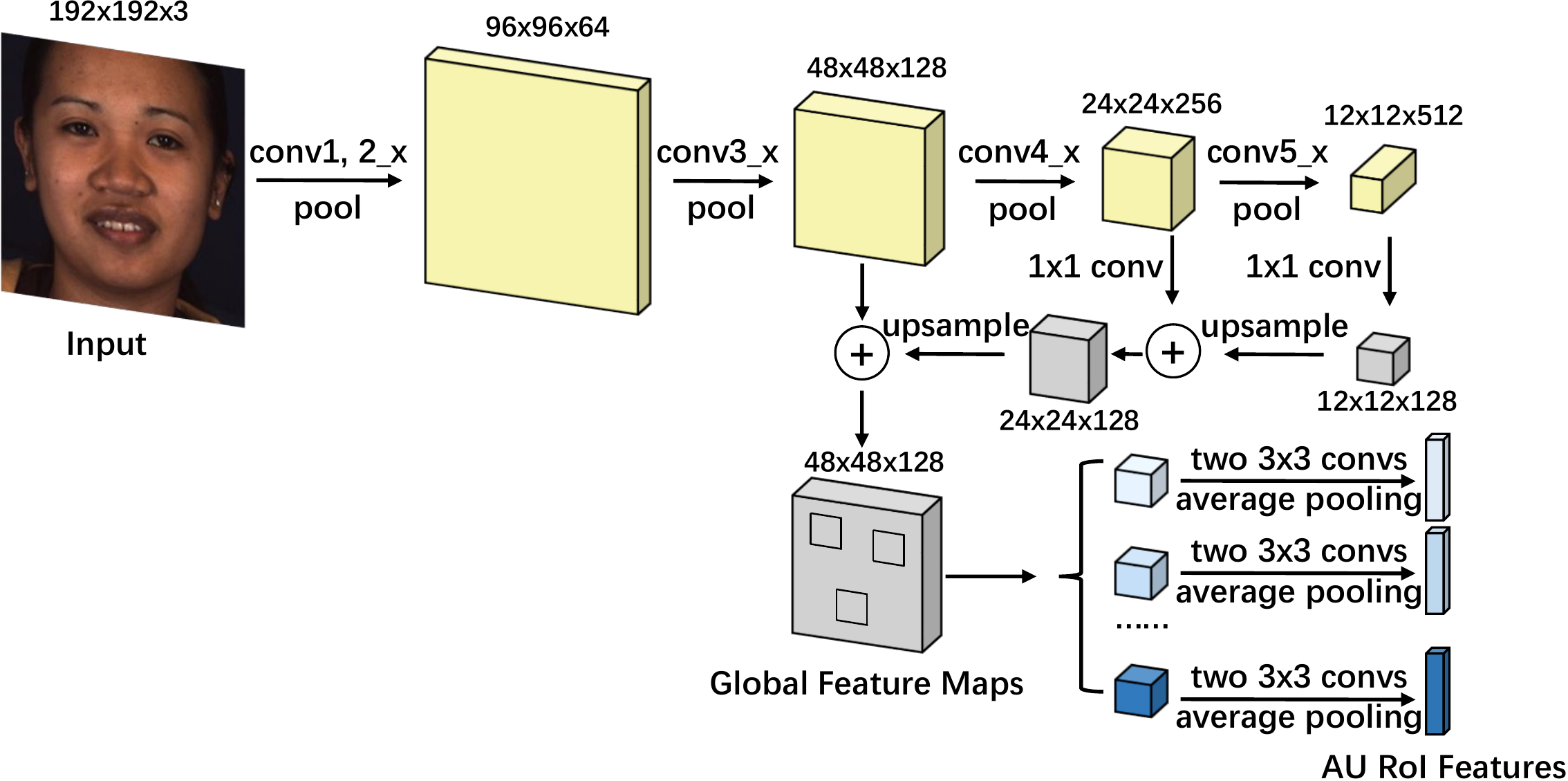}
	\end{center}
	\caption{The network structure of the CNN module in the backbone. \_x denotes for the basic residual blocks. Individual RoI feature learning is applied for each AU separately to extract the RoI features. Best viewed in color.}
	\label{cnn}
\end{figure}

\subsection{Semi-Supervised Learning}
AU recognition based on semi-supervised learning has become very popular recently. Peng et al.~\cite{peng2018weakly} generated pseudo AU labels for images with facial expression labels based on prior AU-expression knowledge and trained the AU classifier in an adversarial fashion. Zhang et al.~\cite{zhang2018classifier} proposed a knowledge-driven method to incorporate the prior knowledge to train multiple AU classifiers. Zhao et al.~\cite{zhao2018learning} proposed a scalable weakly supervised clustering method to refine the noise labels and trained the model based on them. Inspired by traditional co-training, Niu et al.~\cite{niu2019multi} employed two ResNet to generate multi-view features and proposed multi-label co-regularization (MLCR) for AU recognition. Recently by incorporating several dominant methods of semi-supervised learning, MixMatch~\cite{berthelot2019mixmatch} was proposed and achieved superior performance in image classification. However, to the best of our knowledge, there is temporarily no study on integrating semi- and self-supervised learning together for AU recognition.

\section{Method}
In this section, the backbone network is introduced first. Adaptive AU relation embedding and AU related feature extraction via the transformer are presented next. Then two proposed self-supervised tasks, i.e., RoI inpainting and optical flow estimation, are described in detail. Finally, the two tasks are further incorporated with a semi-supervised method to form the unified end-to-end trainable WSRTL framework.

\subsection{Backbone Network}
\begin{figure}[tbp]
	\begin{center}
		\includegraphics[width=0.9\columnwidth]{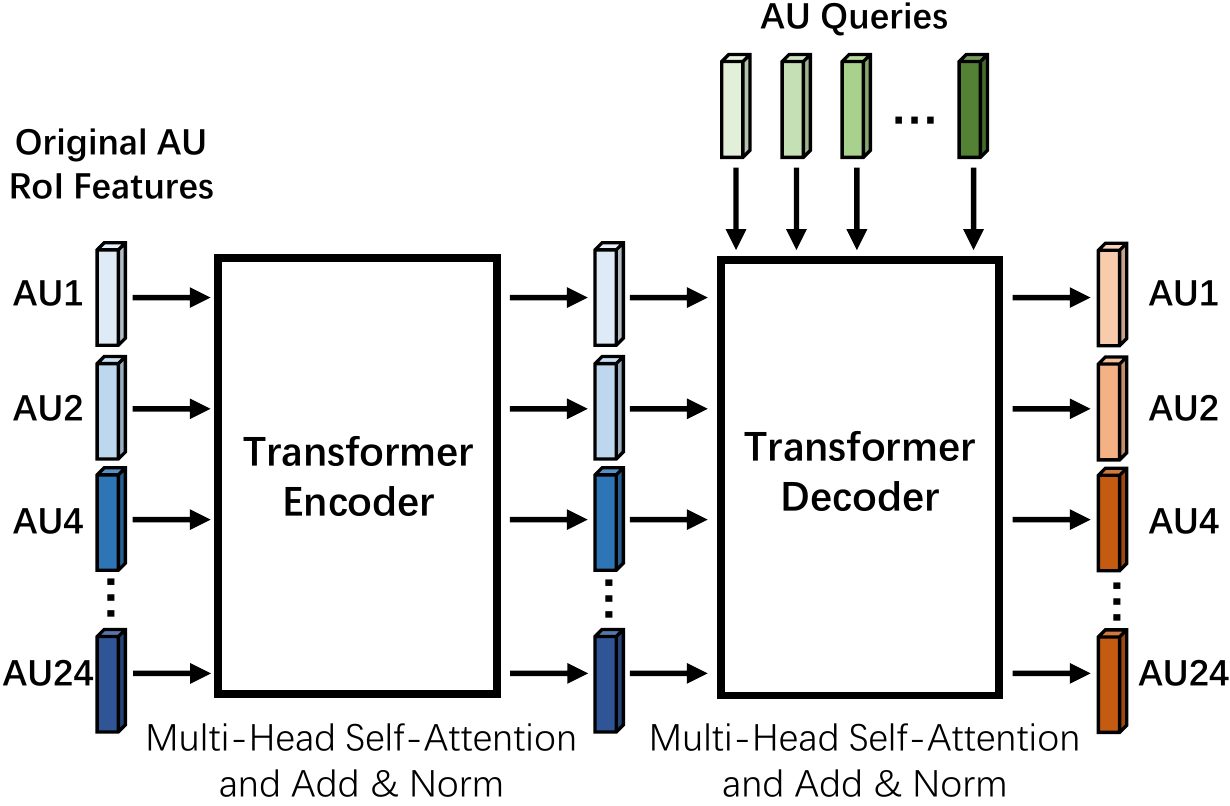}
	\end{center}
	\caption{The transformer takes vectorized RoI features as input and transforms the feature embedding with multi-head self-attention and encoder-decoder attention. AU queries are utilized to extract AU-related features. Best viewed in color.}
	\label{transformer}
\end{figure}

The backbone network consists of three parts, i.e., a conventional CNN module, a transformer and the AU prediction module. The detailed structure of the CNN layers is illustrated in the Figure~\ref{cnn}. ResNet-18~\cite{he2016deep} is chosen as the foundation of the CNN module due to the effectiveness and efficiency in feature learning. Besides that, as each AU is defined on the corresponding local facial region, individual RoI feature learning module which consists of two $3\times 3$ convolutional layers and one average pooling layer for each AU is utilized to deal with different facial textures and structures. We follow the AU RoI definition proposed in~\cite{li2017action}. It is notable that in order to get relatively high spatial resolution feature maps while keeping sufficient semantic information before cropping AU RoIs, the high-level feature maps are gradually upsampled by bilinear interpolation and added with the low-level feature maps as shown in Figure~\ref{cnn}. Then features of each AU RoI are cropped from the fused global feature maps. Due to the symmetry of human face, each AU corresponds to two RoI features. A lightweight transformer, which is composed of one encoder and one decoder, is employed to encode the AU relationship to the RoI features adaptively and further extract AU related features as shown in Figure~\ref{transformer}. We calculate the average of the two enhanced RoI features derived from the transformer for each AU and regard it as the AU RoI representation. Finally, based on local RoI features and overall global features, AU predictions made by separate fully connected layers are fused together by maximizing.

\subsection{Transformer based Adaptive AU Relation Embedding and Feature Extraction}
The transformer was originally designed for nature language processing (NLP) tasks based on the attention mechanism~\cite{vaswani2017attention}. Compared to recurrent neutral networks, it is more parallelizable and easier to train. A typical transformer is composed of multiple encoders and decoders. The feature sequence is first input to the encoder to obtain the attention features adaptively, and then get the corresponding attended features according to the query in the decoder via both self- and encoder-decoder attention.

As a key module in both encoders and decoders, the self-attention layer aims to explore relationship between each input feature and all others in the sequence via the self-attention mechanism. For the transformer encoder in our model, the related AU RoI features can be attended to the corresponding one adaptively and the attention feature can be formulated as follows,
\begin{equation}
  Attention(Q,K,V)=Softmax(\frac{QK^T}{\sqrt{d}})V,
\end{equation}
where $Q$, $K$ and $V$ are linear transformations of the input feature sequence, which is formed as a matrix $X$.  In our model, feature maps obtained from RoI feature learning of each AU are vectorized first by average pooling, and then joined together to form the input feature matrix $X$. Assume there are $N$ AUs and the feature dimension of each RoI is 128, then $X \in \mathbb{R}^ {128\times N}$. After linear transformation, the feature dimension is denoted as $d$. Other modules in the encoder such as add \& norm and feed forward network are the same with the standard structure.

Considering that the propagation process in the encoder mainly focuses on the feature level of each individual image while ignores the generalization of AU features to some extent. Motivated by DETR~\cite{carion2020end}, a transformer decoder is employed to mine more AU related semantic features from the output of the encoder. Specifically, for example, we expect to acquire features closely related to whether AU1 is activated from the regional features by means of a general template or probe which is learned based on the whole database. Such template or probe is named as an AU query.

The learnable AU label embeddings are utilized as queries to explore the essential features for the subsequent AU prediction through the cross-attention mechanism in the decoder. Different from DETR, where the queries are class agnostic, here each query is corresponding to one specific AU, making it more effective to extract AU-related features. Thus, all AU queries can be formulated as a $d\times N$ matrix where each $d$-dimensional vector corresponds to one AU. It is notable that after training, the AU queries should contain the learned AU relation information, which should be similar to the prior AU relation knowledge. We will visualize the encoded relationship in the next section.

It is known that for the transformer based NLP models, position encoding is important for the input feature sequence as the order of words affects the meaning of the sentence. However, regional features of corresponding AUs are already extracted from the convolutional feature maps at fixed locations and spatial information of these AU RoI features are no longer essential since we can directly predict the status of each AU based on corresponding regional features even without a transformer. From another aspect, according to the prior knowledge of AU relationship, the relation of AUs comes from the joint movement of facial muscles instead of the spatial location of AUs. Therefore, positional embeddings are not added in the encoder or the decoder.

The RoI features output from the transformer, which contain enough discriminative and representative information, are then utilized to obtain regional AU predictions and to generate the missing patches for RoI inpainting. Compared to conventional graph propagation methods, where the AU relation graph is often fixed based on prior knowledge or the statistics of AU labels in database, transformer is more flexible and adaptive for each sample. The learning procedure of AU relation encoding and essential feature extraction is totally data driven and does not require to build the AU relation graph manually.

\begin{figure*}[tbp]
	\begin{center}
		\includegraphics[width=0.9\textwidth]{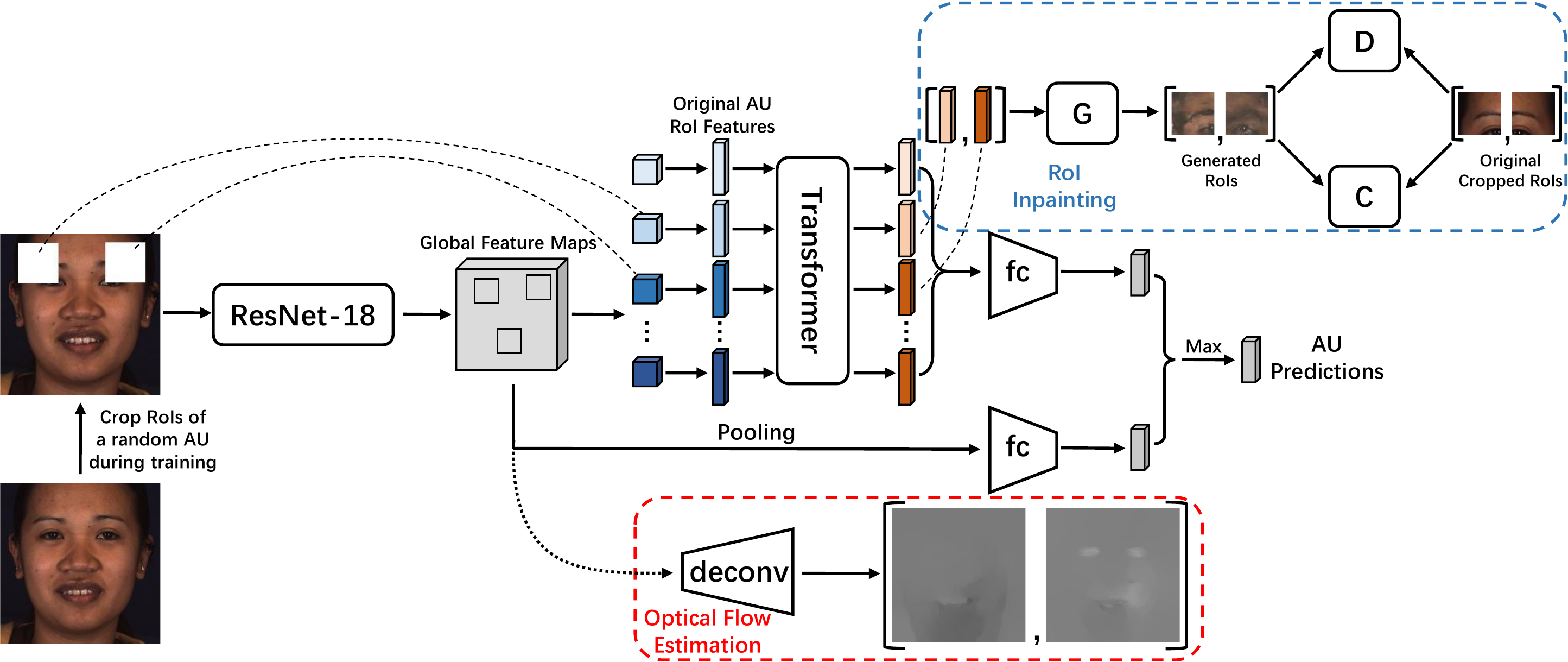}
	\end{center}
	\caption{The proposed WSRTL framework. Backbone network is in the middle. The self-supervised auxiliary tasks of RoI inpainting and optical flow estimation are solved by the modules in the blue and red dashed box, respectively. During the testing phase, only the backbone network is effective and intact facial images are input to the network. Best viewed in color.}
	\label{framework}
\end{figure*}

\subsection{RoI Inpainting}
As AU relationship learned from the limited annotated data may not be accurate and complete, we expect to mine reasonable and comprehensive AU relation knowledge from a large amount of unlabeled data. To this end, we propose a task of RoI inpainting (RoII), which aims for recovering the cropped regions of a random AU in the original facial image. Instead of utilizing global features to generate the cropped patches as conventional image inpainting does~\cite{pathak2016context}, we leverage RoI features of the intact regions and the learned adaptive relationship to represent the cropped parts for recovery to enhance the AU relationship modeling. It is notable that some AUs share the same RoIs, such as AU12 (lip corner puller) and AU15 (lip corner depressor). As they can hardly be activated together, there is no positive correlation between them. When these shared RoIs are cropped, the RoII task can still be completed by the related RoI features.

\subsubsection{Patch Generation with AU Semantic Consistency}
As shown in Figure~\ref{framework}, the symmetrical RoIs of a random AU are first cropped from the original facial image, and then recovered in a generative and adversarial fashion~\cite{goodfellow2014generative} based on RoI features encoded with AU relationship. After the attended RoI feature $x$ of the cropped patch is obtained, we first leverage the vanilla GAN to recover the cropped RoI $p$, where $x\in \mathbb{R}^d$ and $p\in \mathbb{R}^{3 \times s\times s}$, $s$ is the height and width of $p$ in the original image and 3 denotes the RGB channels.

The network module for patch generation is shown in the blue dashed box of Figure~\ref{framework}. The cropped RoIs and the generated patches are treated as real and fake samples respectively. Discriminator $D: \mathbb{R}^{3 \times s\times s}\rightarrow \{1,0\}$ and generator $G: \mathbb{R}^d\rightarrow \mathbb{R}^{3 \times s\times s}$ are trained alternately by the adversarial loss, which is formulated as:
\begin{equation}
  \mathcal{L}_{adv}=\mathbb{E}_{p\sim P}[\log D(p)]+\mathbb{E}_{x\sim X}[\log (1-D(G(x)))],
\end{equation}
where $P=\{ p_1,p_2,...,p_n\}$ and $X=\{ x_1,x_2,...,x_n\}$ are the sets of original cropped patches and attended RoI features derived from the transformer. We train $D$ to maximize $\mathcal{L}_{adv}$ and $G$ to minimize it so that $G$ can finally generate fake samples which cannot be distinguished by $D$. Specifically, the adversarial loss for $G$ can be rewritten as:
\begin{equation}
  \mathcal{L}_{adv}^g=-\mathbb{E}_{x\sim X}[\log D(G(x))].
\end{equation}

Meanwhile, as RoII aims to recover the cropped regions, the reconstruction loss is applied to supervise $G$ during training. Here $l_1$-norm is adopted as follows,
\begin{equation}
  \mathcal{L}_{rec}=\mathbb{E}_{p\sim P, x\sim X}||p-G(x)||_1.
\end{equation}

The vanilla GAN only aims to generate fake facial patches similar to the real ones. Nevertheless, we also expect the RoI features to maintain sufficient AU discrimination besides the appearance information for recovery. Thus an auxiliary AU classifier $C: \mathbb{R}^{3\times s\times s}\rightarrow \{1,0\}$ is employed for the corresponding real and fake patch pairs so that we can explicitly demand the AU semantic of the generated patch is consistent with the original one. Specifically, for unlabeled data, the pseudo label of the cropped AU output from the backbone is regarded as the semantic information. Let $\hat{y}$ denote the pseudo AU label of the cropped patch, then the objective function for the additional AU classifier $C$ is defined as:
\begin{equation}
  \mathcal{L}_C=\mathbb{E}_{p\sim P}[CE(C(p),\hat{y})],
\end{equation}
where $CE$ is the cross-entropy function. When training $G$, $C$ can provide the semantic consistency supervision for generated patches, i.e.,
\begin{equation}
  \mathcal{L}_c^g=\mathbb{E}_{x\sim X}[CE(C(G(x)),\hat{y})].
\end{equation}

Therefore, the final loss functions for $D$ and $G$ are as follows,
\begin{equation}
\begin{split}
  \mathcal{L}_D=&-\mathcal{L}_{adv}\\
  \mathcal{L}_G=&\lambda_1\mathcal{L}_{adv}^g+(1-\lambda_1)\mathcal{L}_{rec}+\lambda_2\mathcal{L}_c^g,
\end{split}
\end{equation}
where $\lambda_1$ and $\lambda_2$ are hyper parameters to balance the losses.

\begin{algorithm*}[tbp]
\caption{The training procedure of weakly supervised regional and temporal learning (WSRTL).}
\label{alg}
\begin{algorithmic}[1]
\REQUIRE Labeled and unlabeled training set $L$ and $U$, number of iterations $m$, batch size $n$, hyper parameters $\lambda_1$, $\lambda_2$, $\lambda_f$ and $\lambda_{u'}$, sharpening temperature $T$
\ENSURE Backbone network $B$, patch generator $G$, patch discriminator $D$, auxiliary AU classifier $C$ and optical flow estimation module $F$
\STATE Load pretrained Imagenet model parameters $\theta_B$ for $B$. Randomly initialize parameters $\theta_G$, $\theta_D$, $\theta_C$ and $\theta_F$ for each module.
\FOR{each $i \in [1,m]$}
\STATE Sample a mini batch of $n$ samples $\{ (l_i, l_i^c, p_i^l, f_i^g, y_i^l)\}_{i=1}^n$ from $L$, where $l_i$ and $y_i^l$ are the image sample and its AU labels, $l_i^c$ is the image after cropping RoIs of a random AU, $p_i^l$ is the cropped RoIs, $f_i^g$ is the corresponding ground truth optical flow.
\STATE Sample a mini batch of $n$ samples $\{ (u_i, u_i^c, p_i^u)\}_{i=1}^n$ from $U$, where $u_i$ is the unlabeled image sample, $u_i^c$ is the image after randomly cropping AU RoIs, $p_i^u$ is the cropped RoIs.
\STATE Calculate the pseudo labels of $\{u_i\}_{i=1}^n$: $y_i^u=\sigma (f_{backbone}(y|u_i;\theta_B)/T)$, where $\sigma$ is the sigmoid function.
\STATE Apply MixMatch to labeled and unlabeled data: $L', U'=MixMatch(L,U)$.
\STATE Update the parameters of $B$ by descending the gradient of $\mathcal{L}_{semi}(l',y'_l,u',y'_u)$ in Eq~\ref{semiloss}.
\STATE Perform the auxiliary tasks of RoII and OFE together with AU recognition.
\STATE Update the parameters of $B$, $D$, $G$, $C$ and $F$ by descending the gradient of $\mathcal{L}(l_i, l_i^c, p_i^l,u_i^c, p_i^u, y_i^l,f_i^g)$ in Eq~\ref{loss}.
\ENDFOR
\end{algorithmic}
\end{algorithm*}

\subsection{Optical Flow Estimation}
In addition to the enhanced discriminative regional features, it is valuable to exploit the motion information of facial muscles for AU recognition. Previous self-supervised learning approaches utilized complex networks to model the transformation from the source frame to the target one to obtain the AU related feature representation. Here we propose a self-supervised task of optical flow estimation based on the static facial image to model the temporal change of facial muscles elegantly and effectively. As optical flow exactly describes the pixel movements of facial muscles between two frames, it can be served as the supervisory signals of the dynamic change of facial textures to guide the network to exploit motion information of corresponding local muscles. 

For a video sequence, we first extract the TV-L1 optical flow~\cite{zach2007duality} between two facial frames. As the frame difference between a very short time step can be negligible while the frame after a long step is not effective for AU recognition in the current frame, the step length is set to 3 frames empirically. Meanwhile, to alleviate the effect of head motion as much as possible, we align the face in the latter frame with the transformation matrix obtained in the previous frame. The extracted optical flow is prepared before training.

As shown in the red dashed box in Figure~\ref{framework}, global feature maps output from ResNet-18 are leveraged to estimate the optical flow. In this way, the backbone network is forced to learn motion related features from the static image. The network structure for optical flow estimation is composed of 2 layers of transposed convolution which aims to gradually model the small pixel-wise displacement. Thus the objective function can be formulated as follows,
\begin{equation}
  \mathcal{L}_F=\mathbb{E}||f_g-f_p||_1,
\end{equation}
where $f_g$ and $f_p$ are the TV-L1 and predicted optical flow respectively. Similar to the reconstruction loss in the previous section, $l_1$-norm is used to measure the difference between the predicted and the TV-L1 optical flow. Another implicit benefit of OFE is that the temporal changes often occur on local facial regions, such as eyes and lip corners, so that the backbone network can also learn to focus on these important regions under this supervision.

\subsection{WSRTL framework}
In traditional self-supervised learning methods, the representation capacity of the trained model via the pretext task is evaluated by training a linear classifier with the labeled data. However, the model performances of previous self-supervised AU recognition methods are not satisfactory~\cite{li2019self,lu2020self}. On the other hand, training the self-supervised task together with the downstream task end-to-end is more efficient and can obtain better results~\cite{zhai2019s4l}. Motivated by this, we incorporate the two auxiliary tasks with the supervised AU recognition task, as illustrated in Figure~\ref{framework}. During training, the auxiliary tasks are also performed on the labeled data. The overall loss function is formulated as:
\begin{equation}\label{loss}
  \mathcal{L}=\mathcal{L}_{Sup}+\mathcal{L}_D+\mathcal{L}_G+\lambda_f\mathcal{L}_F,
\end{equation}
where $\mathcal{L}_{Sup}$ is the binary cross entropy loss function which applied to the labeled data and $\lambda_f$ is the hyper parameter of loss weight for OFE. It is notable that for the image samples with cropped RoIs, we calculate $\mathcal{L}_{Sup}$ only on AUs in the intact regions.

Besides self-supervised learning methods, semi-supervised learning is widely adopted to learn AU related features via unlabeled data. In order to enhance the feature representation from another aspect and maximize the utilization of the unlabeled data, we incorporate the MixMatch~\cite{berthelot2019mixmatch} based semi-supervised learning method in the framework to further boost the model performance.

Unlike the single-label classification problem where softmax is used to get the prediction, AU recognition is a multi-label classification problem and sigmoid is leveraged to output predictions for all AUs. In order to produce lower-entropy predictions for the unlabeled data, the sharpening operation is formulated as follows,
\begin{equation}
  Sharpen(g, T):=\sigma(\frac{g}{T}),
\end{equation}
where $\sigma(\cdot)$ is the sigmoid activation function, sharpening temperature $T\in(0,1]$. 

Assume that after MixMatch, labeled and unlabeled training set $L$ and $U$ are transformed into $L'$ and $U'$. The semi-supervised loss is composed of supervised loss on labeled data and consistency loss on unlabeled data, which can be formulated as follows.
\begin{equation}\label{semiloss}
  \begin{split}
    &\mathcal{L}_{L'}=\mathbb{E}_{(l',y'_l)\sim L'}CE(y'_l, f_{backbone}(l'))\\
    &\mathcal{L}_{U'}=\mathbb{E}_{(u',y'_u)\sim U'}||y'_u-f_{backbone}(u')||^2\\
    &\mathcal{L}_{semi}=\mathcal{L}_{L'}+\lambda_{u'}\mathcal{L}_{U'},
\end{split}
\end{equation}
where $(l',y'_l)$ and $(u',y'_u)$ are the samples and labels of the labeled and unlabeled data from $L'$ and $U'$ respectively. $\lambda_{u'}$ controls the loss weight of the unlabeled data.

Finally, by incorporating the self-supervised auxiliary task learning and semi-supervised learning, the overall WSRTL framework is established. Training procedure for WSRTL is presented in Algorithm~\ref{alg}. In summary, MixMatch is performed firstly and then two self-supervised task are conducted. The parameters of the backbone network are updated twice in each iteration. It is notable that during inference phase, the intact original facial image is input to the model. Only the backbone network is utilized and the modules for two auxiliary tasks are removed.

\section{Experiments}
In this section, ablation studies of the transformer are conducted firstly to validate the effectiveness of the transformer encoder and decoder. Then we make comparison with other self-supervised tasks to prove the superiorities of the two proposed auxiliary tasks. After that, we compare MixMatch with some other semi-supervised learning methods to demonstrate its effectiveness. The integrated framework WSRTL is compared with previous state-of-the-art methods on two frequently used AU recognition benchmarks. The F1-score of single AU and the average F1-score of all AUs in each experiment are reported. Finally, some visualization results are presented and analysed in the last part.

\subsection{Database}
We evaluate our methods on two popular benchmark databases, i.e., BP4D~\cite{zhang2013high} and DISFA~\cite{mavadati2013disfa}. \textbf{BP4D} consists of both 2D and 3D facial expression videos collected from 41 participants, including 23 females and 18 males. Each subject participated in 8 tasks which aimed to induce various facial expressions. There are totally 146,847 frames with valid AU labels. \textbf{DISFA} is another spontaneous AU database which is composed of 27 videos recorded from 12 females and 15 males while they were watching the stimulation material. There are totally 130,815 frames and each frame is manually annotated with AU intensity from 0 to 5. For DISFA database, frames with intensity greater than 1 are treated as positive and the rest are negative. For all experiments, data from these two databases are treated as labeled data and we follow the protocol of subject independent 3-fold cross validation which is widely applied in the community. For unlabeled data, we randomly sample 100,000 images from \textbf{Emotionet}~\cite{fabian2016emotionet} which is a facial expression database containing around one million facial images gathered from the Internet.

\subsection{Evaluation Metric}
AU recognition is a multi-label binary classification problem. As positive and negative samples in most AU databases are very imbalanced, F1-score ($F1=\frac{2PR}{P+R}$), i.e., the harmonic mean of precision and recall, is adopted for fair comparison. In this section, we report the F1-score for each AU and the average F1-score of all AUs, which is denoted as \textbf{Avg.} in the following tables.

\subsection{Implementation Details}
For each frame, we first perform face detection and alignment based on similarity transformation and obtain a $200\times 200$ RGB face image. Ordinary data augmentations such as randomly cropping and horizontally flipping are conducted to enhance the diversity. All images are resized to $192\times 192$ as the input of the network. During testing phase, only center cropping is employed.

The transformer in our model is composed of one layer of encoder and one layer of decoder. For all multi-head self-attention layers, the number of heads is set to 8 empirically. $d$ is set to be 128. Thus, the dimension of each AU query is also 128.

For RoII, we follow~\cite{li2017action} to locate the AU centers and crop a region of $6\times 6$ around each center point from the global feature maps, which corresponds to a $48\times 48$ region on the original image. Facial landmarks are detected with Dlib~\cite{king2009dlib}. $D$ and $C$ are composed of 5 layers of convolution while $G$ consists of 5 layers of transposed convolution. $\lambda_1$, $\lambda_2$ and $\lambda_f$ are set as 0.1, 0.1 and 0.2, respectively. The OFE is addressed with 2 layers of transposed convolution. In MixMatch, the sharpen temperature $T$ and the loss weight of the unlabeled data $\lambda_{u'}$ are empirically set to 0.5 and 1, respectively. ResNet-18 in the backbone takes pre-trained ImageNet model weight as initialization and the rest parameters are initialized randomly. All models are trained via Adam~\cite{kingma2014adam} with the learning rate 0.0003. All experiments are implemented with PyTorch~\cite{paszke2019pytorch} and conducted on one Nvidia Tesla V100.

In the training phase, there are totally 54.62 million parameters and the computational cost is 10.75 GFLOPs. However, during inference, there are only 19.12 million parameters in the network and the computational cost is reduced to 5.57 GFLOPs since the modules working for auxiliary tasks are only effective during training, which validates the computation efficiency in model deploying.

\subsection{Ablation Studies of the Transformer}
\begin{table*}[htbp]
	\centering
    \caption{F1-scores (in \%) on BP4D database for the ablation studies of the transformer. The best and the second best performances are indicated with brackets and bold font, and brackets alone, respectively.}\label{ablation_trans}
	\begin{adjustbox}{max width=\textwidth}
            \begin{tabular}{c|cccccccccccc|c}
            \hline
            AU & 1 & 2 & 4 & 6 & 7 & 10 & 12 & 14 & 15 & 17 & 23 & 24 & \textbf{Avg.}\tabularnewline
            \hline
            GCN & 49.4 & 46.2 & 59.8 & [78.8] & 78.7 & 81.3 & 86.2 & 56.4 & 44.7 & 60.9 & 45.4 & 45.7 & 61.1\tabularnewline
            \hline
            1-layer encoder & 52.8 & 45.7 & [59.4] & 78.6 & [78.7] & 84.4 & 87.5 & 57.6 & [50.7] & 58.9 & 42.2 & 41.4 & 61.5\tabularnewline
            2-layer encoder & [54.2] & 46.4 & 55.9 & 77.8 & 77.2 & [\textbf{84.8}] & 87.0 & 61.2 & 41.5 & 60.9 & 45.0 & 48.5 & 61.7\tabularnewline
            3-layer encoder & [\textbf{54.7}] & 45.2 & 59.3 & 77.9 & 78.0 & 84.1 & [\textbf{87.6}] & 57.5 & 48.6 & [\textbf{63.2}] & 44.9 & 40.0 & 61.8\tabularnewline
            \hline
            Sinusoidal & 47.9 & [47.6] & 57.1 & 78.1 & 77.0 & 84.5 & 87.3 & [\textbf{62.2}] & 49.9 & 56.9 & [47.1] & [50.2] & 62.2\tabularnewline
            Learnable Embedding & 51.0 & [\textbf{48.6}] & [\textbf{61.3}] & 77.7 & 77.7 & [84.6] & 86.7 & 60.1 & 45.6 & [61.4] & 44.8 & [\textbf{50.5}] & [\textbf{62.5}]\tabularnewline
            \hline
            \textbf{Transformer} & 49.2 & 46.3 & 58.7 & [\textbf{79.7}] & [\textbf{78.7}] & 84.2 & [87.5] & [61.5] & [\textbf{51.5}] & 57.8 & [\textbf{47.9}] & 46.8 & [\textbf{62.5}]\tabularnewline
            \hline
            \end{tabular}
	\end{adjustbox}
\end{table*}

Instead of GCN, we propose to utilize a transformer containing of one encoder and one decoder to embed AU relationship adaptively. To demonstrate the superiority of the transformer, we first conduct experiments based on the backbone network with GCN or with a transformer. Here only labeled BP4D samples are used in a fully supervised manner. As shown in Table~\ref{ablation_trans}, compared to GCN, the transformer improves the average F1-score by 1.4\%, which proves its advantage of adaptive AU relation learning and essential feature extraction.

Moreover, in order to validate the effect of the transformer decoder, we compare with the model without a decoder, i.e., only consisting of one layer of encoder. As Table~\ref{ablation_trans} shows, when the decoder is removed, the performance decreases by 1\% and is slightly better than the backbone with GCN. Since the transformer consists of one layer of encoder and one layer of decoder, furthermore, in order to demonstrate the performance gain is not brought by more parameters introduced by the decoder, we design a model with 2 or 3 layers of encoder separately to simulate the increase of parameter number while the decoder is still removed. As presented in the middle of Table~\ref{ablation_trans}, the performance improves slightly as more encoders are added. However, the overall gain is minor in comparison with adding a decoder. These experiments demonstrate that the adaptive AU relation learned by the encoder works better than the fixed prior AU relation and the decoder can mine more discriminative AU features to further improve the performance.

In the proposed model, positional embeddings are not added to the transformer. Nevertheless, we can still conduct experiments with positional embedding involved to verify its effect on AU recognition. Thus, two kinds of positional embeddings are tried to compare with the one without positional embedding. The first is the sinusoidal way~\cite{vaswani2017attention}, and the second is the learnable embedding which is updated during training. The performances on BP4D are presented in Table~\ref{ablation_trans}. When adding the fixed sinusoidal embedding, the performance is slightly worse than the one without it, which proves that the fixed ordinal relation is not beneficial for the regional features based AU recognition. When the learnable position encoding is added, the average F1-score is the same as the one without it, which indicates that the additional position encoding cannot further improve the performance as the encoder itself already exploit the AU relation in a data-driven manner through the self-attention mechanism.

\begin{figure}
	\begin{center}
		\includegraphics[width=\columnwidth]{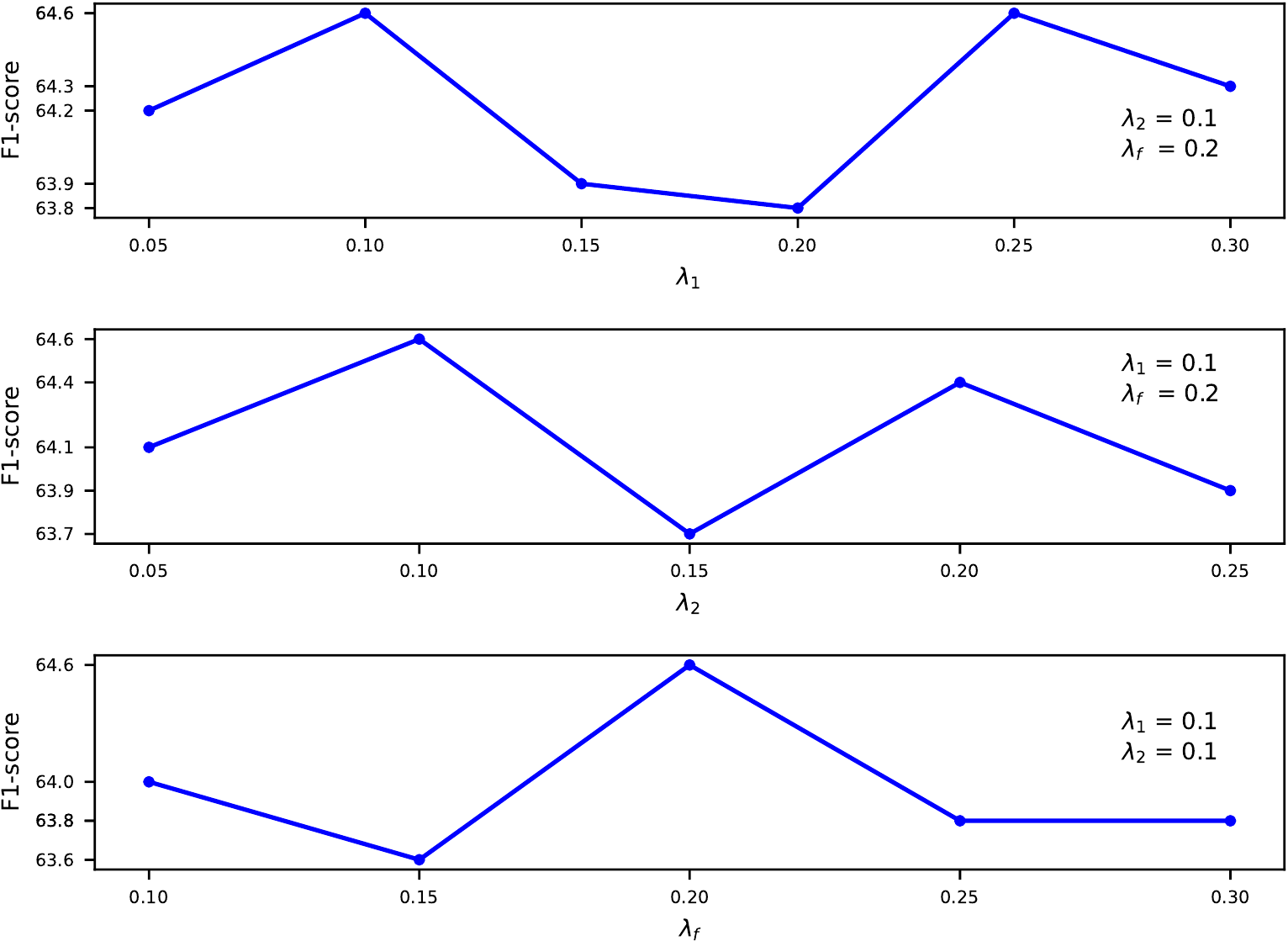}
	\end{center}
	\caption{Model performances (F1-scores, in \%) on BP4D under different settings of $\lambda_1$, $\lambda_2$ and $\lambda_f$.}
	\label{hyperparam}
\end{figure}

The backbone network with a transformer is then regarded as the baseline in the subsequent experiments.

\subsection{Ablation Studies of the Hyperparameters}
For the two self-supervised tasks, we conduct ablation studies for the hyperparameters of $\lambda_1$, $\lambda_2$ and $\lambda_f$ on BP4D. The average F1-scores of 12 AUs under different settings are shown in the Figure~\ref{hyperparam}. As can be seen, the performance fluctuation is within 1\% as the hyperparameters change. When $\lambda_1$, $\lambda_2$ and $\lambda_f$ are set to be 0.1, 0.1 and 0.2, the model achieves the best performance of 64.6\%.

\subsection{Comparison with Other Self-Supervised Auxiliary Tasks}
\begin{table*}[htbp]
	\centering
    \caption{F1-scores (in \%) on BP4D database with different self-supervised tasks. The best and the second best performances are indicated with brackets and bold font, and brackets alone, respectively.}\label{self-supervised-cmp}
	\begin{adjustbox}{max width=\textwidth}
            \begin{tabular}{c|cccccccccccc|c}
            \hline
            AU & 1 & 2 & 4 & 6 & 7 & 10 & 12 & 14 & 15 & 17 & 23 & 24 & \textbf{Avg.}\tabularnewline
            \hline
            Backbone & 49.2 & 46.3 & 58.7 & [79.7] & 78.7 & 84.2 & 87.5 & 61.5 & 51.5 & 57.8 & 47.9 & 46.8 & 62.5\tabularnewline
            \hline
            Rotation & 53.2 & 48.6 & [61.3] & 76.4 & [78.9] & 84.1 & [\textbf{88.4}] & 59.4 & 51.4 & 63.1 & [\textbf{49.0}] & 41.1 & 62.9\tabularnewline
            Exemplar & 54.2 & 45.6 & 55.9 & 77.3 & 77.8 & 83.6 & 87.6 & 63.6 & 49.8 & [64.8] & 46.7 & [52.0] & 63.2\tabularnewline
            Jigsaw & 46.7 & 36.7 & 51.4 & 75.4 & 74.1 & 82.1 & 85.3 & 59.3 & 32.7 & 59.9 & 39.6 & 40.3 & 57.0\tabularnewline
            \hline
            \textbf{OFE} & 49.0 & 46.0 & 57.1 & 77.0 & [\textbf{79.2}] & 83.0 & [88.0] & 64.1 & [52.2] & 64.7 & 50.1 & [\textbf{55.3}] & 63.8\tabularnewline
            \textbf{RoII} & [55.0] & [\textbf{53.0}] & [\textbf{62.8}] & [\textbf{79.8}] & 78.4 & [\textbf{84.6}] & 87.8 & [\textbf{64.5}] & 50.7 & 62.6 & 46.5 & 43.5 & [64.1]\tabularnewline
            \textbf{OFE+RoII} & [\textbf{57.1}] & [49.7] & 60.5 & 77.9 & 76.1 & [84.4] & 87.2 & [64.3] & [\textbf{53.5}] & [\textbf{67.0}] & [48.9] & 48.6 & [\textbf{64.6}]\tabularnewline
            \hline
            \end{tabular}
	\end{adjustbox}
\end{table*}

\begin{table*}[htbp]
	\centering
    \caption{F1-scores (in \%) on BP4D database with different semi-supervised learning methods. The best and the second best performances are indicated with brackets and bold font, and brackets alone, respectively.}\label{semi-comparison}
	\begin{adjustbox}{max width=\textwidth}
            \begin{tabular}{c|cccccccccccc|c}
            \hline
            AU & 1 & 2 & 4 & 6 & 7 & 10 & 12 & 14 & 15 & 17 & 23 & 24 & \textbf{Avg.}\tabularnewline
            \hline
            Backbone & 49.2 & [46.3] & 58.7 & [79.7] & [\textbf{78.7}] & 84.2 & 87.5 & 61.5 & [51.5] & 57.8 & [47.9] & 46.8 & 62.5\tabularnewline
            CutMix & 50.2 & 45.6 & 57.6 & [\textbf{80.3}] & 71.9 & [\textbf{85.8}] & 88.8 & 61.6 & 48.3 & [63.3] & 42.0 & 48.1 & 62.0\tabularnewline
            Mean Teacher & [52.7] & [\textbf{46.9}] & [57.8] & 79.6 & [78.5] & [84.2] & 87.4 & 61.3 & 50.3 & 60.0 & 44.7 & [49.5] & 62.7 \tabularnewline
            FixMatch & 48.8 & 43.9 & 58.6 & 78.8 & 78.2 & 83.2 & [88.8] & [\textbf{64.5}] & [\textbf{51.8}] & [\textbf{64.1}] & 47.0 & [\textbf{49.8}] & [63.1] \tabularnewline
            MixMatch & [\textbf{54.5}] & 45.9 & [\textbf{59.8}] & 79.2 & 77.4 & 84.1 & [\textbf{88.8}] & [64.2] & 50.4 & 62.2 & [\textbf{47.9}] & 46.0 & [\textbf{63.4}] \tabularnewline
            \hline
            \end{tabular}
	\end{adjustbox}
\end{table*}

We conduct experiments to compare the proposed self-supervised auxiliary tasks with several other successful self-supervised tasks in computer vision, including rotation classification~\cite{gidaris2018unsupervised}, exemplar~\cite{dosovitskiy2014discriminative} and jigsaw puzzles~\cite{noroozi2016unsupervised}. Under the same experimental setting, these ones are also served as the auxiliary tasks and jointly trained with AU recognition. Appropriate branches are added to the backbone to solve the corresponding tasks. During testing the branch is removed. For rotation prediction, images are rotated by 0, 90, 180 and 270 degrees. A module of two fc layers is added as the branch to perform the 4-category classification. For exemplar, triplet loss is applied to the global features so that features of different views of the same unlabeled sample are pulled closely and the ones of different samples are pushed away. For jigsaw, we use the concatenated features of the 9 patches to perform jigsaw sequence classification and AU recognition.

All experiments are conducted with the same backbone except for jigsaw puzzle as it requires to divide the input image into 9 patches which is not suitable for RoI feature learning. For this task, only ResNet-18 is employed. The performances are presented in Table~\ref{self-supervised-cmp}. Since the jigsaw task divides the input image into patches which neglects the spatial relations of the joint parts, the model performance is significantly worse than others. After incorporating other self-supervised tasks, AU recognition results are improved to different extents. The rotation and exemplar improve the baseline by 0.4\% and 0.7\% due to better global feature representation.

The proposed OFE task obtains an improvement of 1.3\% in average compared to the baseline and the performance on AU24 is dramatically boosted, which demonstrate the effectiveness of the motion features for certain AUs. The result is also better than examplar as the motion information provides more valuable cues for AU recognition. By enhancing the AU relation learning via RoII, the model achieves the best result among all single self-supervised tasks, which is 64.1\% in average and improves the baseline by 1.6\%. The performance validates the necessity of considering the unique properties of AUs when designing self-supervised tasks. Moreover, when we integrate the two self-supervised auxiliary tasks together and jointly train the model simultaneously, the average F1-score is further improved by 0.5\% compared to RoII alone, as OFE and RoII aim to obtain discriminative AU features from complementary perspectives.

\begin{figure}
	\begin{center}
		\includegraphics[width=0.85\columnwidth]{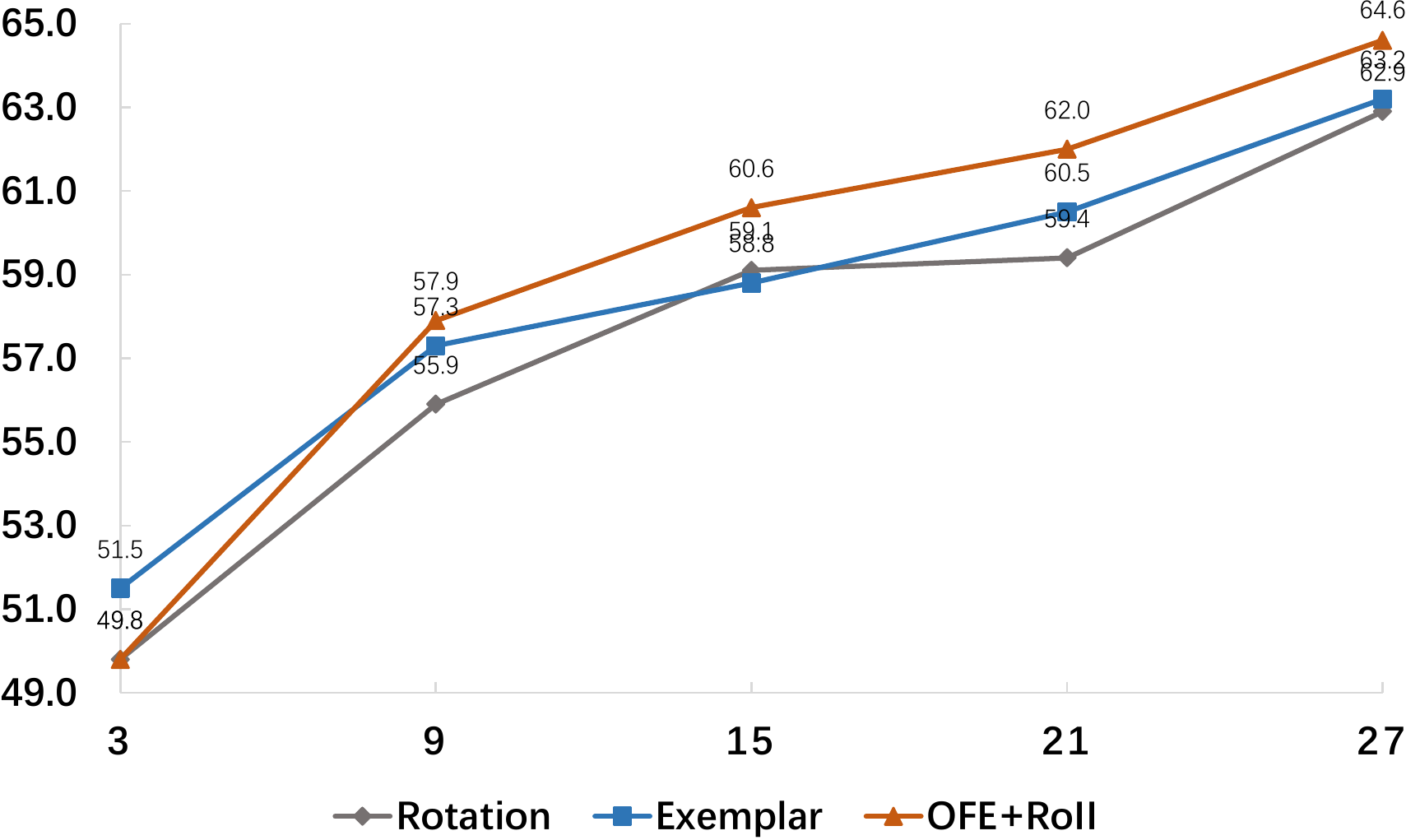}
	\end{center}
	\caption{Model performances (F1-scores, in \%) of different methods trained with an increasing number of labeled subjects, from 3 to 27 in each training set. The unlabeled data used for training is the same.}
	\label{sparse_result}
\end{figure}

Furthermore, to evaluate the model performance when only limited labeled data is accessible, we conduct experiments with a gradually increasing number of labeled training samples. Following the subject independent 3-fold cross validation protocol, instead of using all 27 subjects for training in each fold, we randomly select 3, 9, 15 and 21 subjects and discard the others in the original training set. The unlabeled and validation data are the same with previous experiments. Under the same setting, the performance comparison of three methods are shown in Figure~\ref{sparse_result}. Except for the experiment with only 3 labeled subjects, OFE+RoII outperforms the rotation and exemplar tasks consistently. With around half the original labeled training samples, the OFE+RoII achieves comparable result with the backbone network trained with all training samples, which demonstrates the data efficiency of our method.

\subsection{Experiments on Semi-Supervised Learning}
In this section, we conduct experiments to compare several general semi-supervised learning methods for the task of AU recognition, including CutMix~\cite{yun2019cutmix}, Mean Teacher~\cite{tarvainen2017mean}, FixMatch~\cite{sohn2020fixmatch} and MixMatch~\cite{berthelot2019mixmatch}. The setting of labeled and unlabeled image samples is kept the same as previous experiments and there is an equal number of labeled and unlabeled data in each mini-batch. All methods are preformed with our backbone network.

\begin{table*}[htbp]
	\centering
    \caption{F1-scores (in \%) on BP4D database. The best and the second best performances are indicated with brackets and bold font, and brackets alone, respectively.}\label{bp4d-state-of-the-art}
	\begin{adjustbox}{max width=\textwidth}
            \begin{tabular}{c|cc|cccccc|ccc}
            \hline
            \multirow{2}{*}{AU} & \multicolumn{2}{c|}{Self-Supervised} & \multicolumn{6}{c|}{Supervised} & \multicolumn{3}{c}{Weakly Supervised}\tabularnewline
            \cline{2-12}
            ~ & TCAE & TC-Net & ROI & JAA & DSIN & SRERL & LP-Net & UGN & MLCR & \textbf{OFE+RoII} & \textbf{WSRTL}\tabularnewline
            \hline
            1 & 43.1 & 42.3 & 36.2 & 47.2 & 51.7 & 46.9 & 43.4 & 54.2 & 42.4 & [57.1] & [\textbf{59.7}]\tabularnewline
            2 & 32.2 & 24.3 & 31.6 & 44.0 & 40.4 & 45.3 & 38.0 & 46.4 & 36.9 & [49.7] & [\textbf{51.7}]\tabularnewline
            4 & 44.4 & 44.1 & 43.4 & 54.9 & 56.0 & 55.6 & 54.2 & 56.8 & 48.1 & [60.5] & [\textbf{61.6}]\tabularnewline
            6 & 75.1 & 71.8 & 77.1 & 77.5 & 76.1 & 77.1 & 77.1 & 76.2 & 77.5 & [77.9] & [\textbf{80.3}]\tabularnewline
            7 & 70.5 & 67.8 & 73.7 & 74.6 & 73.5 & [78.4] & 76.7 & 76.7 & 77.6 & 76.1 & [\textbf{80.9}]\tabularnewline
            10 & 80.8 & 77.6 & [85.0] & 84.0 & 79.9 & 83.5 & 83.8 & 82.4 & 83.6 & 84.4 & [\textbf{85.2}]\tabularnewline
            12 & 85.5 & 83.3 & 87.0 & 86.9 & 85.4 & [87.6] & 87.2 & 86.1 & 85.8 & 87.2 & [\textbf{89.7}]\tabularnewline
            14 & 61.8 & 61.2 & 62.6 & 61.9 & 62.7 & 63.9 & 63.3 & [64.7] & 61.0 & 64.3 & [\textbf{67.8}]\tabularnewline
            15 & 34.7 & 31.6 & 45.7 & 43.6 & 37.3 & 52.2 & 45.3 & 51.2 & 43.7 & [53.5] & [\textbf{52.2}]\tabularnewline
            17 & 58.5 & 51.6 & 58.0 & 60.3 & 62.9 & 63.9 & 60.5 & 63.1 & 63.2 & [\textbf{67.0}] & [63.4]\tabularnewline
            23 & 37.2 & 29.8 & 38.3 & 42.7 & 38.8 & 47.1 & 48.1 & 48.5 & 42.1 & [48.9] & [\textbf{51.4}]\tabularnewline
            24 & 48.7 & 38.6 & 37.4 & 41.9 & 41.6 & 53.3 & [54.2] & 53.6 & [\textbf{55.6}] & 48.6 & 46.9\tabularnewline
            \hline
            \textbf{Avg.} & 56.1 & 52.0 & 56.4 & 60.0 & 58.9 & 62.9 & 61.0 & 63.3 & 59.8 & [64.6] & [\textbf{65.9}]\tabularnewline
            \hline
            \end{tabular}
	\end{adjustbox}
\end{table*}

\begin{table*}[htbp]
	\centering
    \caption{F1-scores (in \%) on DISFA database. The best and the second best performances are indicated with brackets and bold font, and brackets alone, respectively.}\label{disfa}
	\begin{adjustbox}{max width=\textwidth}
            \begin{tabular}{c|cc|cccccc|cc}
            \hline
            \multirow{2}{*}{AU} & \multicolumn{2}{c|}{Self-Supervised} & \multicolumn{6}{c|}{Supervised} & \multicolumn{2}{c}{Weakly Supervised}\tabularnewline
            \cline{2-11}
            ~ & TCAE & TC-Net & ROI & JAA & DSIN & SRERL & LP-Net & UGN & \textbf{OFE+RoII} & \textbf{WSRTL}\tabularnewline
            \hline
            1 & 15.1 & 18.7 & 41.5 & 43.7 & 42.4 & 45.7 & 29.9 & 43.3 & [\textbf{57.8}] & [57.3]\tabularnewline
            2 & 15.2 & 27.4 & 26.4 & 46.2 & 39.0 & 47.8 & 24.7 & 48.1 & [\textbf{52.8}] & [51.8]\tabularnewline
            4 & 50.5 & 35.1 & 66.4 & 56.0 & 68.4 & 59.6 & [72.7] & 63.4 & 70.8 & [\textbf{74.3}]\tabularnewline
            6 & 48.7 & 33.6 & [50.7] & 41.4 & 28.6 & 47.1 & 46.8 & 49.5 & [\textbf{53.2}] & 49.8\tabularnewline
            9 & 23.3 & 20.7 & 8.5 & 44.7 & 46.8 & 45.6 & [49.6] & 48.2 & [\textbf{52.7}] & 44.8\tabularnewline
            12 & 72.1 & 67.5 & [\textbf{89.3}] & 69.6 & 70.8 & 73.5 & 72.9 & 72.9 & 74.5 & [79.3]\tabularnewline
            25 & 82.1 & 68.0 & 88.9 & 88.3 & 90.4 & 84.3 & [93.8] & 90.8 & 91.5 & [\textbf{94.6}]\tabularnewline
            26 & 52.9 & 43.8 & 15.6 & 58.4 & 42.2 & 43.6 & [\textbf{65.0}] & 59.0 & 51.9 & [64.6]\tabularnewline
            \hline
            \textbf{Avg.} & 45.0 & 39.4 & 48.5 & 56.0 & 53.6 & 55.9 & 56.9 & 60.0 & [63.1] & [\textbf{64.6}]\tabularnewline
            \hline
            \end{tabular}
	\end{adjustbox}
\end{table*}

The results are presented in Table~\ref{semi-comparison}. The performance of the fully supervised backbone network is in the first line. Motivated by RoII, we cut the RoIs of a random AU from a unlabeled image and then paste them to the corresponding locations of a labeled image. The consistency loss is applied to make the prediction for the cut AU of the cutmix one is consistent with the original unlabeled image. However, the cutmix method performs even worse than baseline due to that the cutmix operation disrupts the AU relation learning to some extent. For Mean teacher, the parameters of teacher model are updated by the exponential moving average (EMA) of previous student models. It performs slightly better than baseline. For FixMatch, the reliable pseudo label of weakly augmented unlabeled data is utilized as the supervisory signal for the strong augmentation of the same image. Thus the model can learn better feature representation through large amounts of unlabeled data. FixMatch improves the baseline by 0.6\% which indicates the effectiveness of the pseudo label for unlabeled data. By means of the integration of mixup~\cite{zhang2017mixup}, entropy minimization~\cite{grandvalet2005semi}, consistency regularization~\cite{laine2016temporal} and EMA, MixMatch performs best among these semi-supervised methods and achieves 63.4\% in average F1-score. Therefore MixMatch is selected for the WSRTL framework.

\subsection{Comparison with State-of-the-art Methods}
We compare WSRTL with previous state-of-the-art methods on BP4D and DISFA, including the supervised methods such as ROI~\cite{li2017action}, JAA-Net~\cite{shao2018deep}, DSIN~\cite{corneanu2018deep}, SRERL~\cite{li2019semantic}, LP-Net~\cite{niu2019local}, UGN~\cite{song2021uncertain} and the self-supervised methods such as TCAE~\cite{li2019self} and TC-Net~\cite{lu2020self}.

Experimental results on BP4D are presented in Table~\ref{bp4d-state-of-the-art}. Obviously there is a performance gap between previous self-supervised approaches and our method. It is because the two-stage training framework does not take advantage of the labeled AU data to train the backbone network. Meanwhile, previous self-supervised methods mainly focus on global feature learning while our self-supervised tasks are more AU specific, which incorporates regional and temporal feature learning, and AU relation encoding. Compared to the fully supervised models, WSRTL fully explores the unlabeled data to capture better feature representation via the auxiliary self-supervised tasks and semi-supervised learning, resulting in an improved performance. In comparison with other auxiliary task, such as face alignment in JAA, ours does not require additional annotations but gets a better result. Meanwhile, compared to the semi-supervised method MLCR~\cite{niu2019multi} which also leverages the unlabeled data, WSRTL significantly outperforms it by 6.1\% and sets a new state-of-the-art performance.

Table~\ref{disfa} presents the experimental results on DISFA. Compared to the previous state-of-the-art supervised method UGN, WSRTL achieves an improvement of 4.6\% and obtains the overall performance of 64.6\%. In comparison with BP4D, DISFA contains less training samples with more severe data imbalance problem. Most of the previous methods suffer from significant performance degradation. However, in our method, by fully leveraging large amounts of unlabeled data, WSRTL maintains competitive performances on majority of the AUs consistently and achieves the new state-of-the-art performance on DISFA.

\subsection{Visualization Analysis}

\begin{figure*}[tbp]
	\begin{center}
		\includegraphics[width=0.8\textwidth]{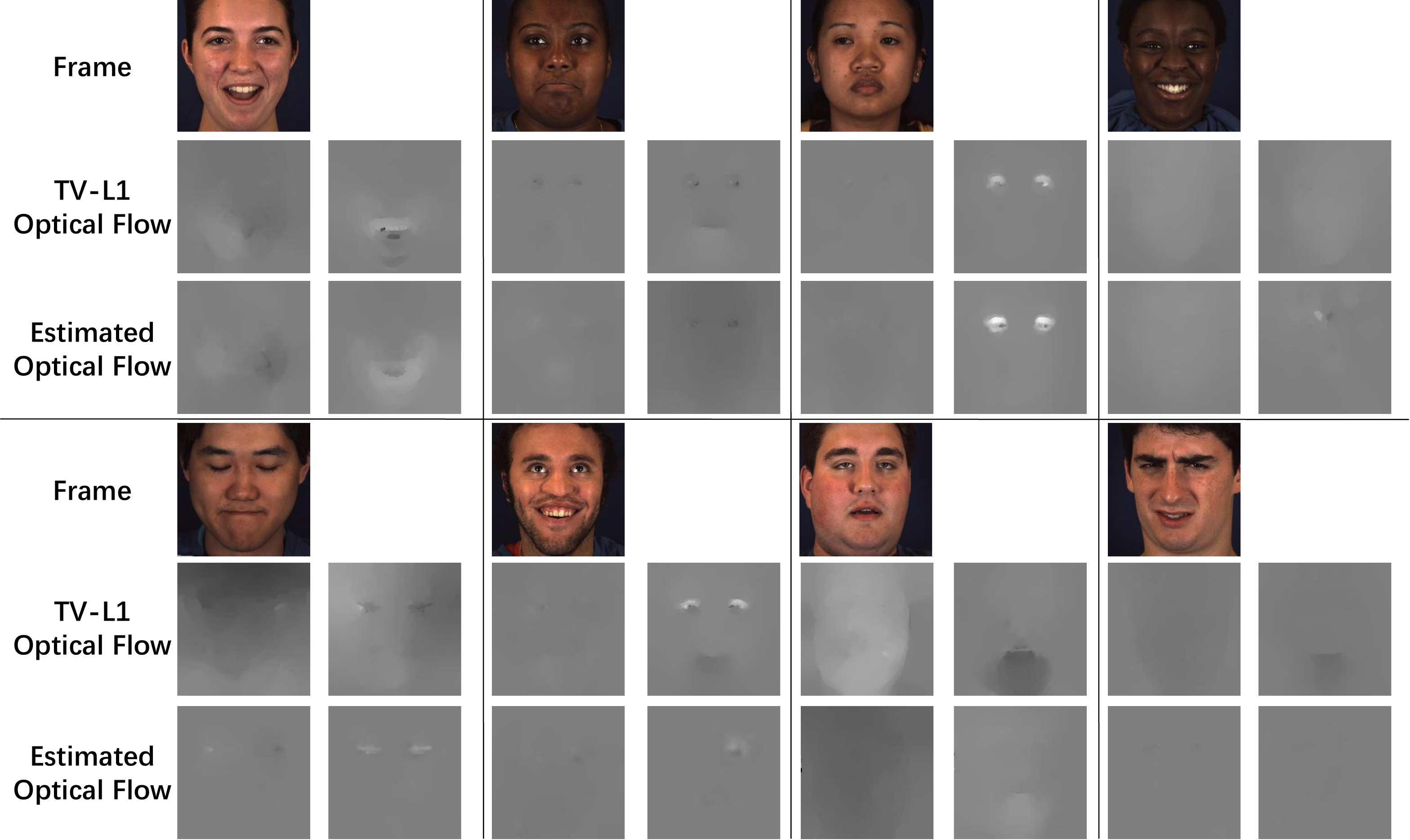}
	\end{center}
	\caption{Visualization for optical flow estimation on BP4D samples. The gray images show the optical flow in the horizontal and vertical directions respectively.}
	\label{flow}
\end{figure*}

\begin{figure}[tbp]
	\begin{center}
		\includegraphics[width=0.9\columnwidth]{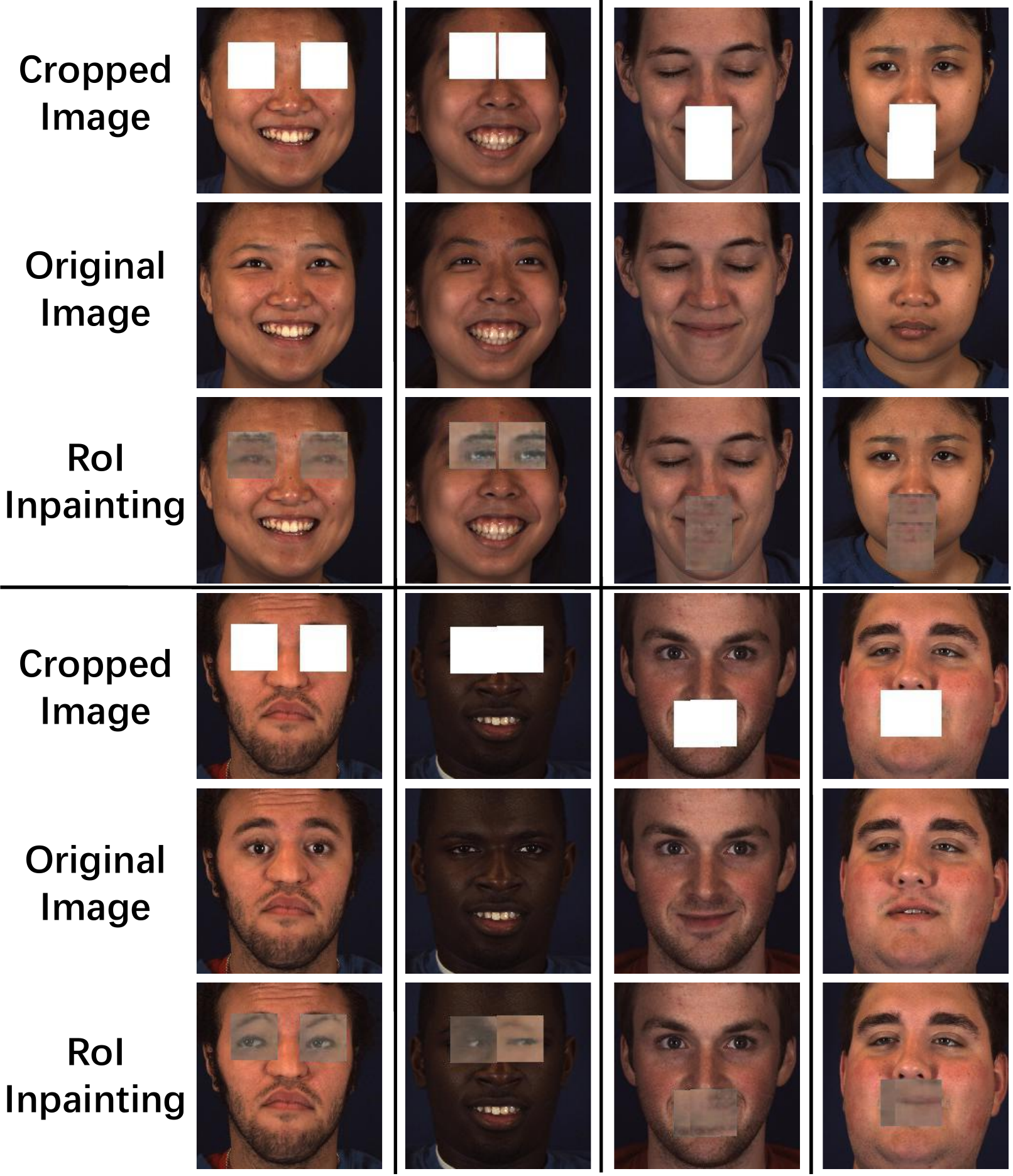}
	\end{center}
	\caption{Visualization for RoI inpainting on BP4D samples. Best viewed in color.}
	\label{inpainting}
\end{figure}

The estimation of optical flow for 6 subjects in BP4D is illustrated in Figure~\ref{flow}. We use grayscale to visualize the optical flow in horizontal and vertical directions respectively, where $I_x$ is shown in left and $I_y$ is in right. TV-L1 optical flow extracted between two frames is treated as ground truth for the task of single image based OFE. As can be seen, optical flow describes the motion information of important facial regions where an AU is activated. The predictions are shown in the third row. Most of the predictions are similar to the ground truth which indicates the learned temporal features encoded in the global features are effective. Thus, through the OFE task, the backbone network can extract temporal features from single image to further enhance the discrimination of global features for AU recognition.

The visualization for RoI inpainting is shown in Figure~\ref{inpainting}. In the first row, symmetrical RoIs of a random AU are removed and replaced with white color. The original images are shown below. The third row presents the results of RoI inpainting. From the aspect of facial semantic, most of the regions around eyes and lips are recovered correctly which verifies the effectiveness of RoI features embedded with adaptive AU relation via the transformer. By the auxiliary task of RoI inpainting, the representation and discrimination of regional features are enhanced. However, as the dimension of regional feature used for recovery is only 128, and the network module for patch generation utilized is not sophisticated, some generated regions may seem poor in appearance and suffer from misalignment, e.g., some left and right eyes are not distinguishable. Nevertheless, the proposed RoII task aims to better explore the AU relation adaptively and embed it to the regional feature representation rather than pursue great facial appearance details in the color space. In other words, the muscle movements of the cropped region derived from other intact regions via AU relation embedding matter more in this task. From this perspective, the RoII works well to meet our demand.

\begin{figure}
	\centering
	\subfigure[BP4D]
	{
		\begin{minipage}{0.46\columnwidth}
   			\includegraphics[width=\columnwidth]{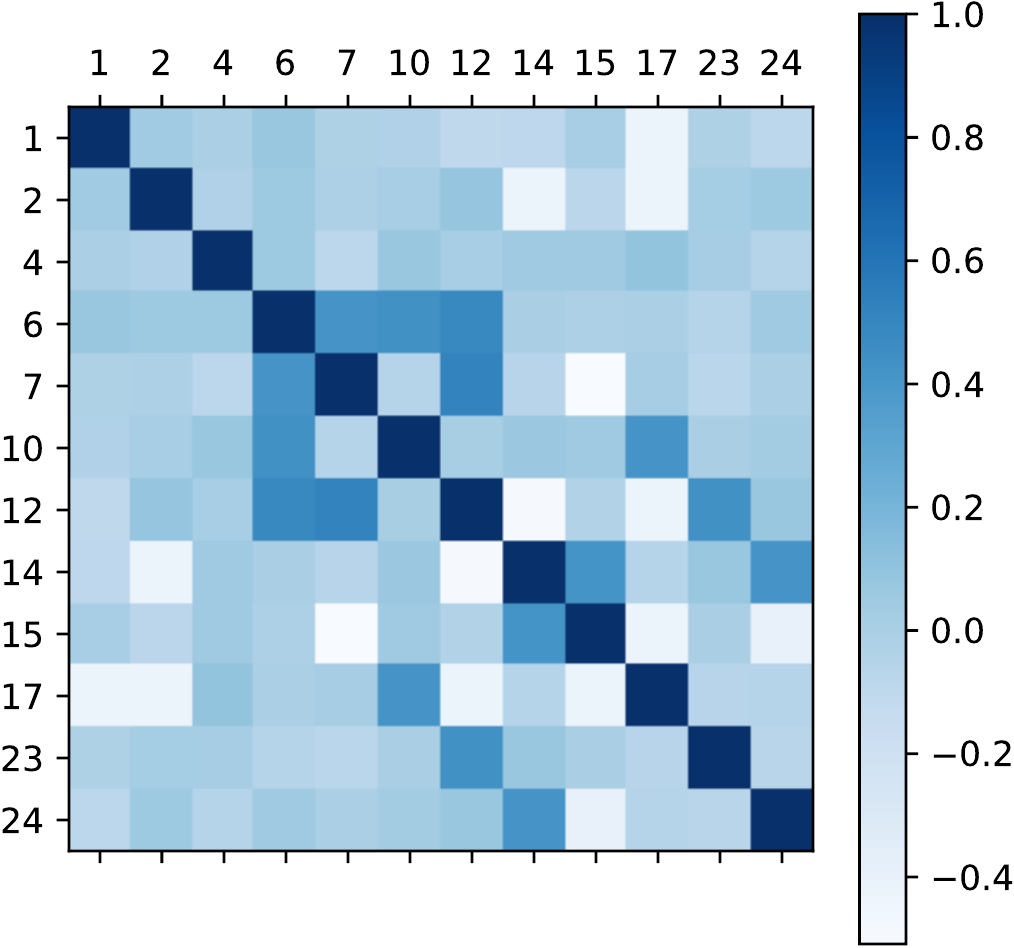}
		\end{minipage}
	}
	\subfigure[DISFA]
	{
		\begin{minipage}{0.46\columnwidth}
			\includegraphics[width=\columnwidth]{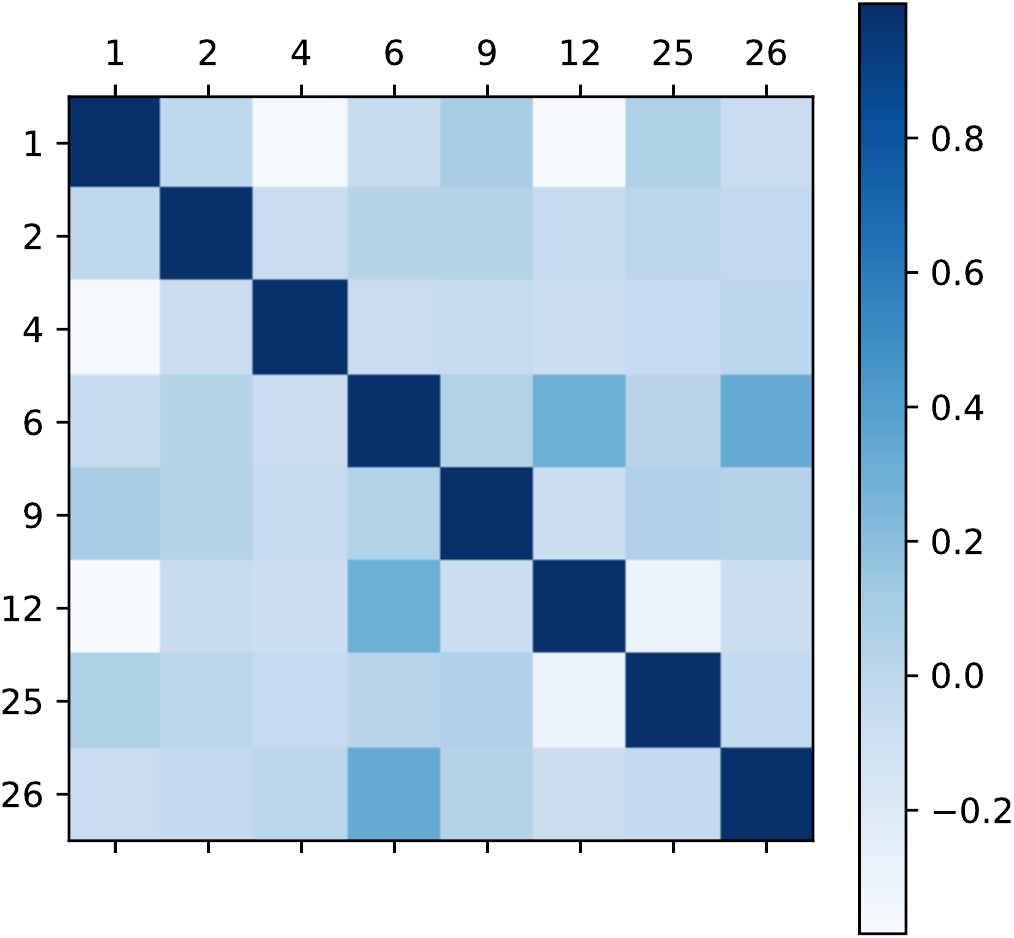}
		\end{minipage}
	}
	\caption{Cosine similarity matrices of AU queries in the transformer decoders for BP4D and DISFA respectively, which indicate the learned relationship knowledge between AUs. Best viewed in color.}
	\label{cos_sim}
\end{figure}

As elaborated in Section III.B, AU queries in the transformer decoder should contain AU relation knowledge implicitly after training. In order to visualize the learned AU relation, we calculate the cosine similarity between each AU query and show the similarity matrix with heatmap in Figure~\ref{cos_sim}. The relation of two AUs is closer if the blue color of the corresponding region is darker. Most of the relationships presented in the Figure~\ref{cos_sim} are consistent with the facial action coding expertise~\cite{peng2018weakly}, e.g., AU10 (upper lip raiser) and AU17 (chin raiser) are related in BP4D, and AU6 (cheek raiser and lid compressor), AU12 and AU26 (jaw drop) are closely related in DISFA. Moreover, the similarity matrix of BP4D also reveals some unique relations in the database, e.g., there exists close relationship between AU6, AU7 (lid tightener), AU10 and AU12. It is because AU6 and AU7 are very similar, and the annotators of BP4D tend to annotate AU10 as positive when the subject is laughing which is reasonable. Therefore, the AU queries in our model successfully captures a general AU relation knowledge from the whole database.

\section{Conclusion}
In this paper, motivated by the unique properties of AUs, we take the regional and temporal feature learning, together with AU relation embedding into consideration, and design two novel auxiliary tasks in a self-supervised manner to leverage large amounts of unlabeled data to improve the performance of AU recognition. Meanwhile, a transformer is utilized to embed the AU relation adaptively and extract more discriminative regional features in the backbone network. Furthermore, we incorporate the two self-supervised tasks with semi-supervised learning, and propose the joint framework WSRTL, which is end-to-end trainable and can capture more discriminative features from labeled and unlabeled data. Extensive experimental results demonstrate the superiority of our method and new state-of-the-art performances on BP4D and DISFA have been achieved.

\bibliographystyle{IEEEtran}
\bibliography{mybibfile}

\begin{IEEEbiography}[{\includegraphics[width=1in,height=1.25in,clip,keepaspectratio]{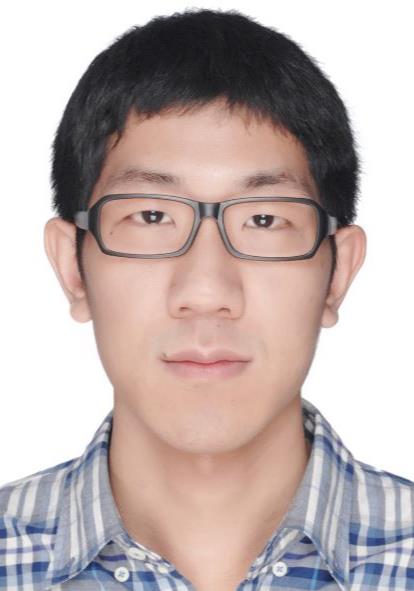}}]{Jingwei~Yan}
received the B.S. and Ph.D. degrees from Southeast University, Nanjing, China, in 2011 and 2019, respectively. He has been with the Hikvision Research Institute, Hangzhou, since 2019. His current interests include affective computing, computer vision and pattern recognition.
\end{IEEEbiography}

\begin{IEEEbiography}[{\includegraphics[width=1in,height=1.25in,clip,keepaspectratio]{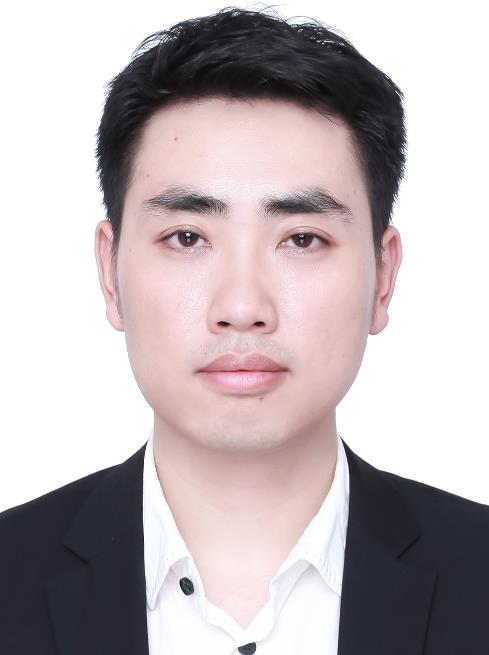}}]{Jingjing~Wang}
received the B.S. and Ph.D. degrees from the Department of Electronic Engineering and Information Science, University of Science and Technology of China (USTC), China, in 2010 and 2016, respectively. He is currently with the Hikvision Research Institute, Hangzhou. His research interests include computer vision and machine learning.
\end{IEEEbiography}

\begin{IEEEbiography}[{\includegraphics[width=1in,height=1.25in,clip,keepaspectratio]{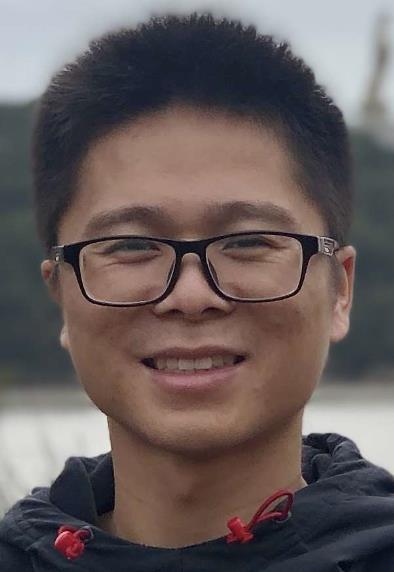}}]{Qiang~Li}
received the B.S. degree from Nanjing Normal University, Nanjing, in 2014, and the M.S. degree from Southeast University, Nanjing, in 2017. He is currently an algorithm engineer at Hikvision Research Institute, Hangzhou, China. His research interests include computer vision and machine learning.
\end{IEEEbiography}

\begin{IEEEbiography}[{\includegraphics[width=1in,height=1.25in,clip,keepaspectratio]{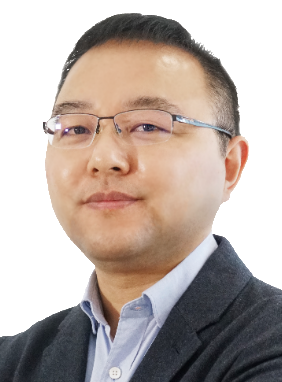}}]{Chunmao~Wang}
received the B.S. and M.S. degrees from Southeast University, Nanjing, China, in 2008 and 2011, respectively. He is currently the director of the Biometrics Research Department at the Hikvision Research Institute. His current interests include computer vision and machine learning.
\end{IEEEbiography}

\begin{IEEEbiography}[{\includegraphics[width=1in,height=1.25in,clip,keepaspectratio]{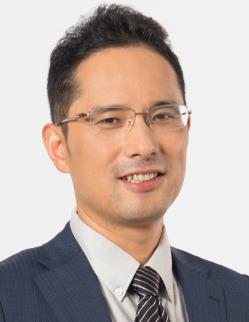}}]{Shiliang~Pu}
received the Ph.D. degrees from the University of Rouen Normandy and Zhejiang University in 2005 and 2007. He is currently a Chief Research Scientist at Hikvision, and the President of the Hikvision Research Institute. His research interests include AI, machine perception, and robotics.
\end{IEEEbiography}


%




\end{document}